\documentclass{article}
\PassOptionsToPackage{numbers}{natbib}
\usepackage[sglblindworkshop, final]{neurips_2025}
\workshoptitle{SimBioChem 2025}

\usepackage[utf8]{inputenc} % allow utf-8 input
\usepackage[T1]{fontenc}    % use 8-bit T1 fonts
\usepackage{hyperref}       % hyperlinks
\usepackage{url}            % simple URL typesetting
\usepackage{booktabs}       % professional-quality tables
\usepackage{amsfonts}       % blackboard math symbols
\usepackage{nicefrac}       % compact symbols for 1/2, etc.
\usepackage{microtype}      % microtypography
\usepackage{xcolor}         % colors

\usepackage{amsmath}
\usepackage{amssymb}
\usepackage{amsthm}
\usepackage{bbold}
\usepackage{mathtools}
\usepackage{listings}
\usepackage{microtype}
\usepackage{graphicx}
\usepackage{booktabs} 
\usepackage{hyperref}
\usepackage{xurl}
\usepackage{algorithm}
\usepackage{caption}
\usepackage{subcaption}
\usepackage{wrapfig}

\usepackage{latexsym}

\usepackage[textsize=tiny]{todonotes}
\title{Sparse Data Diffusion\\for Scientific Simulations in Biology and Physics}

\author{%
    Phil Sidney Ostheimer\textsuperscript{1}, Mayank Nagda\textsuperscript{1}, Andriy Balinskyy\textsuperscript{1}, Jean Radig\textsuperscript{2},\\ \textbf{Carl Herrmann\textsuperscript{2}, Stephan Mandt\textsuperscript{3}, Marius Kloft\textsuperscript{1}, Sophie Fellenz\textsuperscript{1}}\\
    \textsuperscript{1}RPTU University Kaiserslautern-Landau, \textsuperscript{2}Heidelberg University, \textsuperscript{3}University of California, Irvine\\
    \texttt{phil.ostheimer@rptu.de}\\
}
\begin{document}

\maketitle

\begin{abstract}
Sparse data is fundamental to scientific simulations in biology and physics, from single-cell gene expression to particle calorimetry, where exact zeros encode physical absence rather than weak signal. However, existing diffusion models lack the physical rigor to faithfully represent this sparsity. This work introduces Sparse Data Diffusion (SDD), a generative method that explicitly models exact zeros via Sparsity Bits, unifying efficient ML generation with physically grounded sparsity handling. Empirical validation in particle physics and single-cell biology demonstrates that SDD achieves higher fidelity than baseline methods in capturing sparse patterns critical for scientific analysis, advancing scalable and physically faithful simulation.
\end{abstract}

\section{Introduction}

Sparse data---i.e., data where most values are zero---is fundamental to scientific simulations in biology and physics. Sparsity arises naturally from physical principles: in calorimeter images from high-energy experiments, energy deposits occur only in localized regions corresponding to particle interactions; in scRNA data, each cell expresses only a subset of genes  \citep{Lu:2019,Lu:2021,Lahnemann:2020}. 

Formally, we focus on data that is continuous yet exhibits a discrete sparsity pattern:
\begin{equation*}
\mathbf{x} \in \mathbb{R}^d, \; \text{where} \; |\mathbf{x}|_0 \ll d.
\end{equation*}
That is, most entries of the vector $\mathbf{x}$ with continuous values are exactly zero.

In scientific contexts, exact zeros encode physical meaning---a silent gene represents biological absence, not weak expression; an empty calorimeter cell indicates no particle interaction, not low energy. This physical interpretation makes sparsity a fundamental constraint rather than merely a statistical pattern. As \citet{Donoho:2006} noted in compressed sensing, most data ``can be thrown away'', and \citet{Tibshirani:1996} emphasized with the Lasso that sparsity reflects reality: only a small subset of variables is truly relevant.

\begin{figure}[ht]
\centering
\begin{subfigure}[b]{0.49\columnwidth}
    \centering
    \includegraphics[width=\linewidth]{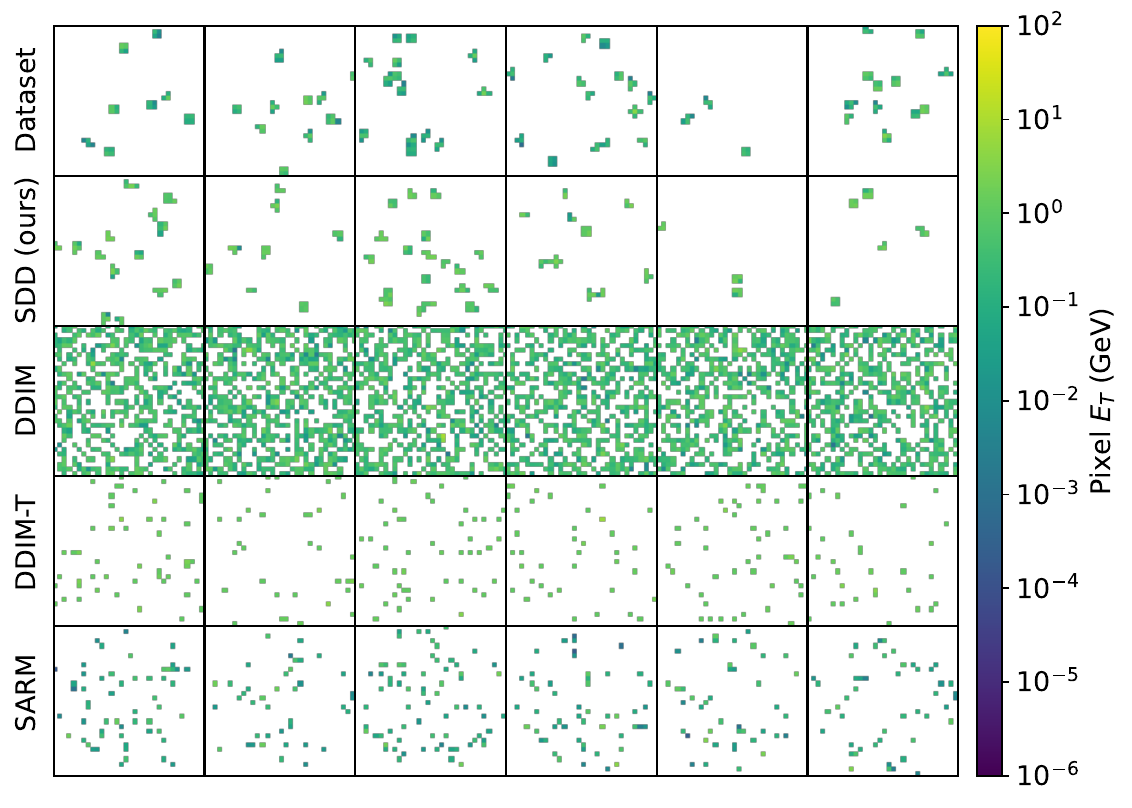}
    \caption{Muon Signal}
    \label{fig:ddim_muon_signal}
\end{subfigure}
\begin{subfigure}[b]{0.49\columnwidth}
    \centering
    \includegraphics[width=\linewidth]{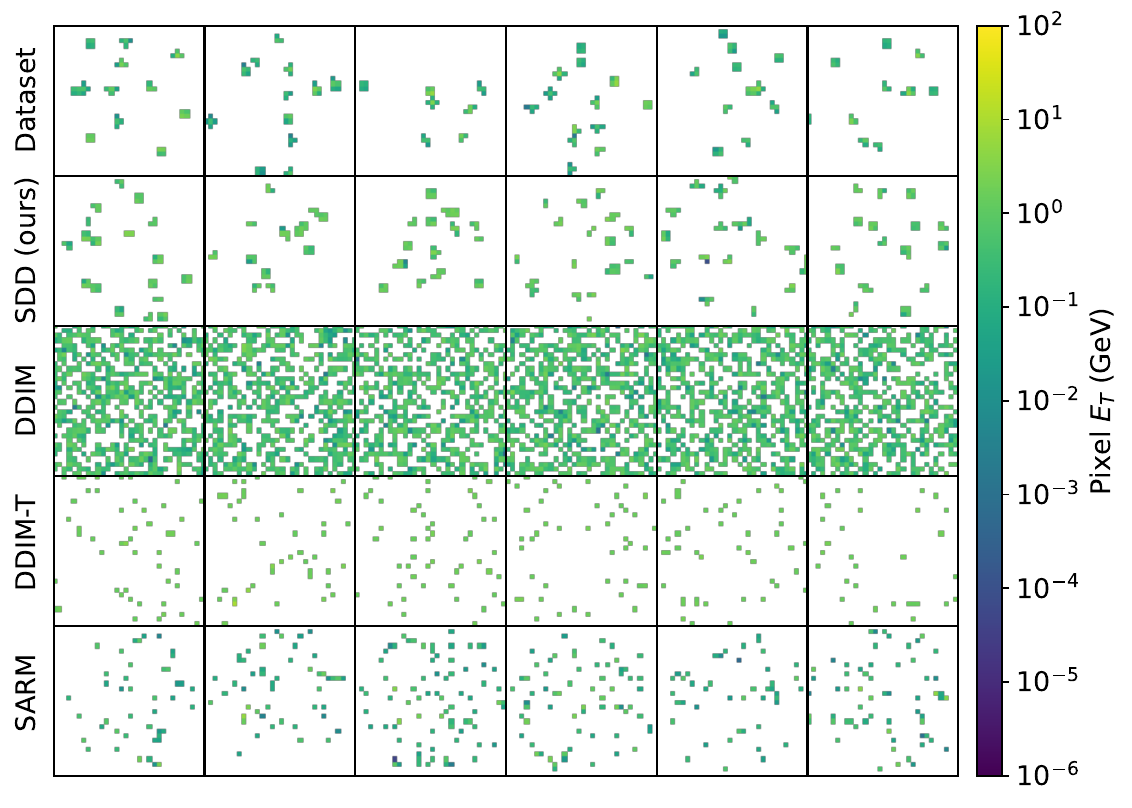}
    \caption{Muon Background}
    \label{fig:ddim_muon_background}
\end{subfigure}
\caption{Calorimeter images from the muon isolation study: signal images (left) and background images (right). Rows show samples from (1) real data, (2) SDD (ours), (3) DDIM, (4) DDIM with post-hoc thresholding to match dataset sparsity (DDIM-T), and (5) the domain-specific SARM baseline. Pixel intensity (GeV) visualizes energy deposition per cell (white=zero). SDD uniquely recovers the distinct, sparse, and clustered patterns characteristic of real data, whereas DDIM completely misses the sparsity, and DDIM-T and SARM show unrealistic isolated energy deposits.}
\label{fig:ddim_muon}
\end{figure}

While machine learning has accelerated scientific simulation, generative models---including Generative Adversarial Networks (GANs) \citep{Goodfellow:2020}, Variational Autoencoders (VAEs) \citep{Kingma:2014}, and Diffusion Models (DMs) \citep{Sohl-Dickstein:2015,Ho:2020,Song:2020,Austin:2021}---fundamentally mismatch physically sparse data. Adding Gaussian noise to every dimension assumes smooth, recoverable perturbations, but true zeros encode meaningful physical absence and become indistinguishable from weak signal once noised. Their isotropic noise processes, smooth denoising networks, and activations (ReLU \citep{Nair:2010}, Tanh, Sigmoid \citep{Lecun:2012}) bias outputs toward density, compromising physical fidelity. DDIM produces only 49\% sparsity on data that is 95\% sparse (Fig.~\ref{fig:ddim_muon}), and thresholding (DDIM-T) to match dataset sparsity still fails to recover the clustered structures observed in real particle physics data. This underscores the need for generative models that treat sparsity as a physical constraint rather than a post-hoc correction.

To bridge this gap, we introduce Sparse Data Diffusion (SDD), a physically-grounded generative framework that explicitly models sparsity as a fundamental constraint. At its core, SDD uses \emph{Sparsity Bits} (SBs)---discrete latent variables that indicate, for each output dimension, whether it should be active or zero based on the underlying physical structure. Dense values are modeled using a continuous state-space diffusion process, preserving scalability while maintaining physical fidelity. During sampling, SBs enforce exact zeros, ensuring generated data respects the physical constraints of scientific data. We validate SDD primarily on challenging scientific datasets from particle physics and single-cell biology, with additional evaluation on computer vision benchmarks to demonstrate generality. In summary, we make the following contributions:
\begin{itemize}
\item We introduce SDD, a physically-grounded DM that bridges scalable ML with physical accuracy by explicitly modeling sparsity as a physical constraint in DMs.\footnote{Code: \url{https://github.com/PhilSid/sparse-data-diffusion}}
\item We propose Sparsity Bits---discrete latent variables that encode physical structure---and integrate them with continuous DMs for dense values, enabling joint modeling.
\item We validate SDD on scientific datasets from physics and biology, achieving substantial improvements over standard DMs and domain-specific baselines while maintaining physical fidelity, with additional demonstrations on computer vision tasks.
\end{itemize}

\section{Related work}
As no broadly applicable generative model for sparse data exists, our section on related work summarizes existing domain-specific works on sparse data generation, related concepts in diffusion models, and enforcing sparsity in neural model parameters (and not the data itself).

\paragraph{Sparse data generation}
Past work on sparse, \emph{continuous} data generation focuses on simple architectures and specific applications, such as calorimeter sensor data in physics, which are difficult to compare to general-purpose generative models as they incorporate lots of domain-specific knowledge. They decouple the sparsity information for generating data by introducing decoupled generative models where the sparsity is introduced as a learnable Dirac delta mass at zero \citep{Lu:2019,Lu:2021}. Other works focusing on sparse,  \emph{discrete} data generation use Poisson or negative binomial distributions to model sparse count data with bursts \citep{Zhou:2012,Schein:2016}. There are also deep variants \citep{Gong:2017,Guo:2018, Schein:2019}. However, as already indicated, they do not apply to our setting, which has continuous sparse data that do not exhibit the traits of a Poisson or negative binomial distribution.

\paragraph{Diffusion models}
Diffusion models can be broadly categorized into continuous state-space models \citep{Sohl-Dickstein:2015,Ho:2020,Song:2020,Song:2021,Pandey:2023} for continuous data and discrete state-space models \citep{Austin:2021,Gu:2022} for discrete data. While discrete models can represent sparse discrete data exactly, they are not applicable to our sparse continuous setting. Some continuous models, such as Bit Diffusion \citep{Chen:2023}, perform strongly on discrete data, and others like Unigs \citep{Qi:2024} and DefectSpectrum \citep{Yang:2024} handle mixed variable types by first mapping discrete variables into continuous embeddings. Recent work has also explored diffusion under unconventional data characteristics, including heavy-tailed distributions \citep{Pandey:2025}. However, these approaches do not jointly diffuse continuous and discrete variables, whereas our method enables their simultaneous diffusion within a unified framework.

\paragraph{Enforcing sparsity} Sparsity is not just an inherent trait in datasets that can be, e.g., exploited to store data efficiently. Sparsity can also help in the model architecture. Previous work \citep{Han:2015,Ullrich:2017} shows that neural networks are greatly overparameterized and multiple methods \citep{Molchanov:2017,Louizos:2018,Sun:2024,Lee:2018} inter alia have been introduced to mitigate this overparameterization by sparsifying the underlying neural network weights. Others \citep{Hu:2022} leverage this property to efficiently train low-rank decomposition matrices, which are added as sparse weight updates to the existing model weights. However, these methods focus on enforcing sparsity in the model weights and are therefore not applicable to our setting, where we enforce sparsity in the model output.

\section{Method}
\label{sec:method}
This section presents Sparse Data Diffusion (SDD), a novel framework designed for the generation of sparse data. We begin by introducing the underlying statistical model, followed by an in-depth explanation of the forward and backward diffusion processes, and the overall training and sampling procedure.

\subsection{Statistical model} 
The following statistical model forms the basis for SDD: Let $\mathbf{x_0} \sim p$ with $\mathbb{E}\left[\mathbf{x_{0}}\right]<\infty$ and $\mathbf{x_0} \in \mathbb{R}^d$, where $p$ is unknown and arbitrary. We infer the sparsity for each dimension by applying the indicator function element-wise to $\mathbf{x_0}$:
\begin{equation}
    \mathbf{\bar{x}_0}=2*\mathbb{1}_{x\neq0}(\mathbf{x_0}) -1
\end{equation}

As a result, $\mathbf{\bar{x}_0} \in \{-1,1\}^d$ is a binary vector that encodes, for each element in $\mathbf{x_0}$, whether it is zero or represents a dense value. Therefore, we refer to each element in $\mathbf{\bar{x}_0}$ as a \emph{Sparsity Bit (SB)}. We obtain the extended input $\mathbf{\hat{x}}_0 \in \mathbb{R}^{2d}$, where $\mathbf{\hat{x}}_0 \sim \hat{p}$, by concatenating $\mathbf{x_0} \in \mathbb{R}^{d}$ and $\mathbf{\bar{x}}$:
\begin{equation}
    %\mathbf{\hat{x}_0}=\langle \mathbf{x_0},\mathbf{\bar{x}_0} \rangle
    \mathbf{\hat{x}_0} = \begin{bmatrix} \mathbf{x_0} \\ \mathbf{\bar{x}_0} \end{bmatrix}
\end{equation}

Unlike prior methods, our forward–backward process diffuses continuous and discrete variables jointly. Previous work operated either only on continuous variables \citep{Sohl-Dickstein:2015,Ho:2020,Song:2020,Song:2021} or on discrete variables \citep{Chen:2023}. 

\paragraph{Forward diffusion} The forward diffusion process follows previous work \citep{Sohl-Dickstein:2015,Ho:2020,Song:2020,Song:2021}. It consists of a predefined series of transitions from the input space $\mathbf{\hat{x}_0} \in \mathbb{R}^{2d}$ to pure noise $\boldsymbol{\epsilon} \in \mathbb{R}^{2d}$, where $\boldsymbol{\epsilon} \sim \mathcal{N}(0,I)$. The transition from $\mathbf{\hat{x}_0}$ to $\mathbf{\hat{x}_t}$ is defined as
\begin{equation}
    \mathbf{\hat{x}_t} = \sqrt{\alpha(t)}\mathbf{\hat{x}_0} + \sqrt{1-\alpha(t)}\boldsymbol{\epsilon},
\end{equation}

where $\boldsymbol{\epsilon} \sim \mathcal{N}(0,I)$, $t \sim \mathcal{U}(0,T)$ is a continuous time variable, and $\alpha$ is the noise schedule, a monotonically decreasing function from 1 to 0. In the limit, we obtain:
\begin{equation}
    \lim_{T\to\infty} \mathbf{\hat{x}_T} = \boldsymbol{\epsilon} \sim \mathcal{N}(0,I),
\end{equation}

\begin{figure}[t!]
\begin{center}
\centerline{\includegraphics[width=\columnwidth]{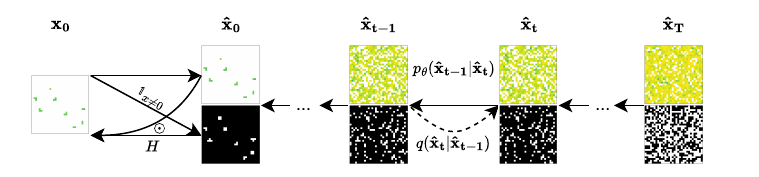}}
\caption{Shown is an illustration of our method SDD. Compared to other diffusion models, we expand the continuous input by discrete Sparsity Bits for forward diffusion and use them in backward diffusion and sampling to enforce sparsity in the data.}
\label{fig:graphical_model}
\end{center}
\end{figure}

\paragraph{Backward diffusion} The backward diffusion process consists of steps that reverse the forward diffusion process:   $\mathbf{\hat{x}_T} \rightarrow \mathbf{\hat{x}_{T-\Delta}} \rightarrow ... \rightarrow \mathbf{\hat{x}_0}$. These steps follow a normal distribution:
\begin{equation}
    \mathbf{\hat{x}_{t-1}} \vert \mathbf{\hat{x}_{t}} \sim \mathcal{N}(\mu_{t}(\mathbf{\hat{x}_t},\mathbf{\hat{x}_0}),\sigma_{t}^2)
\end{equation}
where $\mathbf{\hat{x}_0}$ is the denoised input. Since $\mathbf{\hat{x}_0}$ is not given in the backward diffusion process, we train
a denoising neural network $f_{\theta}$ to predict $\mathbf{\hat{x}_0}$. We describe the training process in the following Section \ref{sec:training} and illustrate our model in Figure \ref{fig:graphical_model}. 

\subsection{Training}
\label{sec:training}
In order to perform the steps $\mathbf{\hat{x}_T} \rightarrow \mathbf{\hat{x}_{T-\Delta}} \rightarrow ... \rightarrow \mathbf{\hat{x}_0}$ of the backward diffusion process, we train a denoising network $f_{\theta}$ to predict $\mathbf{\hat{x}_0}$, which is part of the distribution $p(\mathbf{\hat{x}_{t-1}} \vert \mathbf{\hat{x}_{t}})$. The predicted value of $\mathbf{\hat{x}_0}$ is approximated by  $\mathbf{\tilde{x}_0}=f_{\theta}(\mathbf{\hat{x}_t},t)$  (alternative formulations predict the noise $\boldsymbol{\epsilon}$). 

We use self-conditioning \citep{Chen:2023} to incorporate the previously computed $\mathbf{\tilde{x}_0}$ (or $\boldsymbol{\tilde{\epsilon}}$) from time step $t+1$ to compute the next $\mathbf{\tilde{x}_0}$ (or $\tilde{\epsilon}$).  The training is realized using an $l_2$ regression loss on the continuous inputs and a cross entropy loss (CE on labels in $\{-1,1\}$) on SBs:
\begin{equation}
\label{eq:loss}
\begin{aligned}
    \mathcal{L}(\theta) = 
    \mathbb{E}\Big[ & \left\Vert f_{\theta,0:d-1}(\mathbf{\hat{x}_t},t,\mathbf{\tilde{x}_0})-\mathbf{\hat{x}_{0,0:d-1}}\right\Vert^2 \\
    & + \mbox{CE}\left(f_{\theta,d:2d-1}(\mathbf{\hat{x}_t},t,\mathbf{\tilde{x}_0}),\mathbf{\hat{x}_{0,d:2d-1}}\right) \Big],
\end{aligned}
\end{equation}
where $\mathbf{\hat{x}_0} \sim \hat{p},t \sim \mathcal{U}(0,T),\boldsymbol{\epsilon
} \sim \mathcal{N}(0,I)$ and $\mathbf{x_t}=\sqrt{\alpha(t)}\mathbf{\hat{x}_{0}}+\sqrt{1-\alpha(t)}\boldsymbol{\epsilon}$. We summarize the SDD training process in Algorithm \ref{alg:training} in Appendix \ref{app:train_sample}.

\subsection{Sampling} To draw samples, we use the same state transitions as described in the backward diffusion process: $\mathbf{\hat{x}_T} \rightarrow \mathbf{\hat{x}_{T-\Delta}} \rightarrow ... \rightarrow \mathbf{\hat{x}_0}$, following $p(\mathbf{\hat{x}_{t-1}} \vert \mathbf{\hat{x}_{t}})$ using $f_{\theta}(\mathbf{\hat{x}_t},t,\mathbf{\tilde{x}_0})$. There are multiple ways to perform these steps. Here, we focus on Denoising Diffusion Probabilistic Models (DDPM) \citep{Ho:2020} and Denoising Diffusion Implicit Models (DDIM) \citep{Song:2020}. 

We perform one final step: $\mathbf{\hat{x}_0} \rightarrow \mathbf{x_0}$. Since the first $d$ dimensions in $\mathbf{\hat{x}_0}$ contain  the dense and second $d$ dimensions in $\mathbf{\hat{x}_0}$ contain the SBs, we quantize the SB output dimensions $\mathbf{\hat{x}_{0;d:(2d-1)}}$ using the heaviside step function $H$ applied element-wise to extract the SBs. Afterward, we apply an element-wise product to get the sparsified output:
\begin{equation}
    \mathbf{x_0} = \mathbf{\hat{x}_{0;0:(d-1)}} \odot H(\mathbf{\hat{x}_{0;d:(2d-1)})}
\end{equation} 
We summarize the sampling algorithm in Algorithm \ref{alg:sampling} in Appendix \ref{app:train_sample}.

\section{Experiments}
To evaluate SDD for sparse data generation, we first assess the overall sparsity of generated data, then report results on challenging scientific applications in physics and biology, and conclude with an analysis of sparse image generation.

\subsection{Experimental setup}
\label{sec:experimental_setup}
\paragraph{Baselines} 
We compare SDD to two well-known diffusion models—DDPM \citep{Ho:2020} and DDIM \citep{Song:2020}—which are designed for dense data generation. We provide architectural and training details in Appendix \ref{app:dm_architecture}. To evaluate the performance on sparse outputs, we create thresholded variants DDPM-T and DDIM-T that apply iteratively increasing thresholds (linearly spaced between zero and maximum output value) to match training data sparsity. This provides a pragmatic baseline for adapting diffusion models to sparse domains without explicit sparsity modeling during training.

For particle physics, we benchmark against SARM D+C \cite{Lu:2021}, a domain-specific model that autoregressively samples calorimeter data using spiral patterns informed by inner and outer regions. For scRNA, we include scDiffusion \cite{Luo:2024}, which uses an autoencoder pretrained on a large cell corpus but lacks explicit sparsity enforcement.

\paragraph{Datasets} 
In physics, we use calorimeter images from a muon isolation study \citep{Lu:2019}, represented as 32$\times$32 pixel grids with 95\% sparsity. Pixel intensity represents the transverse momentum ($P_T$) deposited in each cell. The dataset contains 33,331 signal images of isolated muons and 30,783 background images of muons with jets. For biology, we use two scRNA datasets: Tabula Muris \citep{Schaum:2018} with 57K cells (90\% sparsity, 98\% when filtered to 1000 highly variable genes) and Human Lung Pulmonary Fibrosis \citep{Habermann:2020} with 114K cells (91\% sparsity, 96\% filtered). For vision, we use MNIST \citep{Lecun:1998} (60K images, 81\% sparsity) and Fashion-MNIST \citep{Xiao:2017} (60K images, 50\% sparsity), both 28$\times$28 grayscale. All datasets are linearly scaled to $[-1, 1]$ following \citet{Ho:2020}, with inverse transformation applied after inference.

\paragraph{Evaluation}
We evaluate sparsity distribution matching and use domain-specific metrics. For physics, we compute the normalized Wasserstein distance $W_P$ between distributions of transverse momentum $P_T$ and invariant mass for the dataset and 50,000 generated images \citep{Lu:2019,Lu:2021}. For scRNA, following \citep{Luo:2024}, we use Spearman Correlation (SCC), Pearson Correlation (PCC), Maximum Mean Discrepancy (MMD) \citep{Gretton:2012}, and Local Inverse Simpson's index (LISI) \citep{Haghverdi:2018}, comparing real data to 10,000 generated cells. For images, we measure sparsity after discretizing logit values from $[-1,1]$ to $[0,255]$ and compute Fr\'echet Inception Distance (FID) \citep{Heusel:2017} between 50,000 generated images and training data. Note that FID's validity is limited for sparse data as the Inception network was trained on dense ImageNet images \citep{Deng:2009}.

\subsection{Evaluating the recovery of ground truth sparsity patterns}
\label{sec:sparsity_characteristics}

\begin{figure}[ht]
\includegraphics[width=\columnwidth]{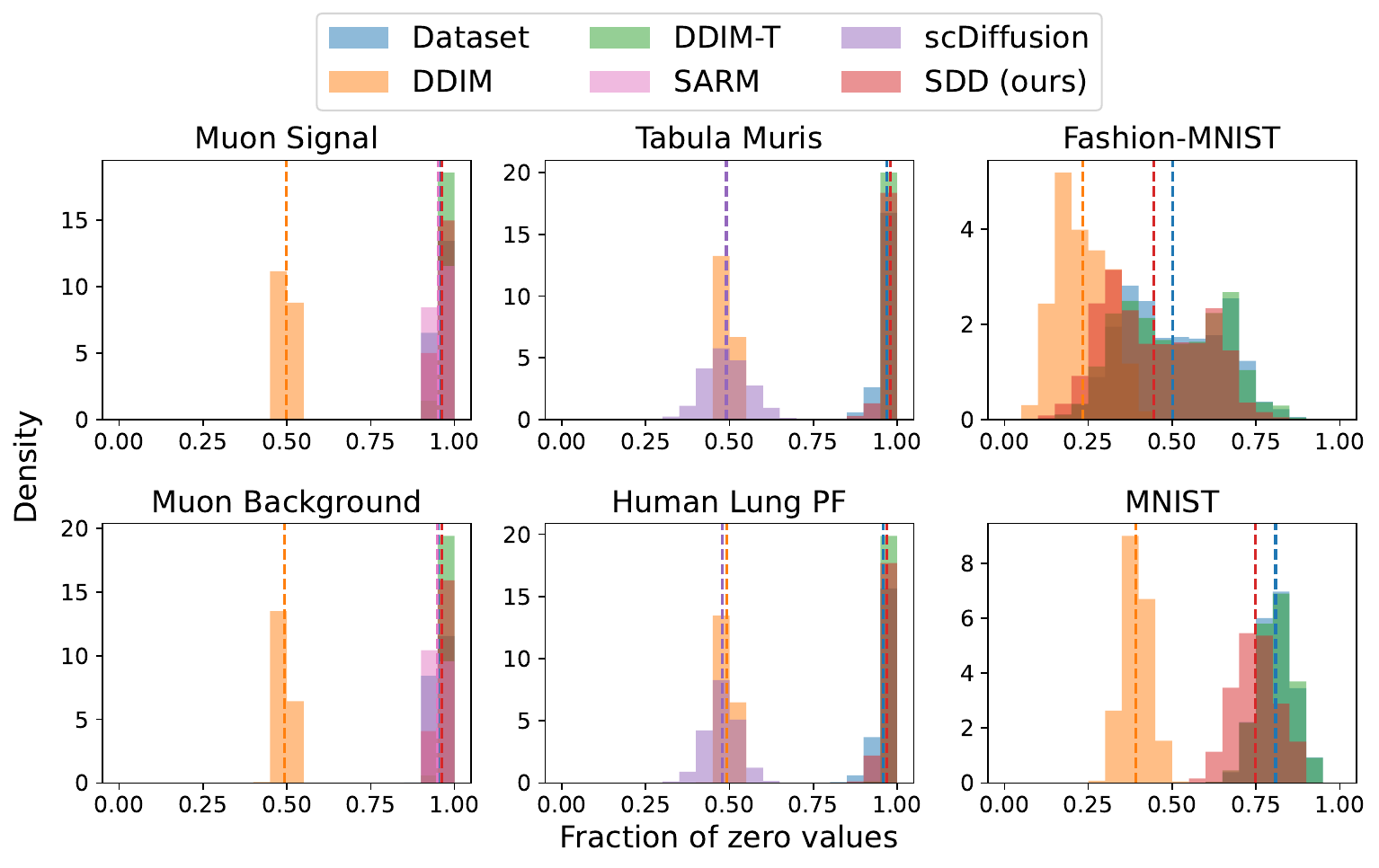}
\caption{Sparsity distribution of real and generated data. Histograms (20 bins) and average sparsity levels (dashed lines) compare real data to samples from DDIM, DDIM-T, SARM, scDiffusion, and SDD. DDIM and scDiffusion underestimate sparsity; DDIM-T matches the average but lacks diversity and overshoots at high sparsity. SARM also underestimates sparsity, while SDD accurately matches both average value and distribution.}
\label{fig:ddim_sparsity_distribution}
\end{figure}

\paragraph{How well do models match ground-truth sparsity?} 
We analyze the sparsity distribution of generated samples in Figure~\ref{fig:ddim_sparsity_distribution} for DDIM, with DDPM results in Appendix \ref{app:further_sparsity_analysis} largely consistent with DDIM. Across all datasets, DDIM and the domain-specific scDiffusion tend to underestimate sparsity, producing outputs denser than the real data. While DDIM-T matches average sparsity by design, it often overshoots in the upper range and truncates the lower end of the distribution, especially for sparse image datasets. The particle physics-specific SARM also slightly underestimates sparsity despite domain knowledge. In contrast, SDD stands out by accurately capturing both the average sparsity and the overall sparsity distribution, achieving realistic and faithful sparsity patterns without relying on domain-specific knowledge or post-hoc processing.

In general, diffusion models struggle to generate sparse data. If the standard post-processing step—clipping values to the range $[-1, 1]$ followed by rescaling—is omitted, the resulting samples exhibit near-zero sparsity across all datasets. As this clipping step is standard practice in diffusion-based models, we retain it and focus our subsequent analyses on clipped outputs.

\paragraph{Consequences of sparsity mismatch in scientific and vision tasks} 
In \emph{particle physics}, identifying particle sources from detector signatures is crucial \citep{Lu:2019,Lu:2021}. DDIM and DDIM-T fail to reproduce data sparsity (Figure~\ref{fig:ddim_muon}), producing overly dense outputs that inflate particle detections and lead to incorrect interpretations. In \emph{scRNA data}, sparsity reflects biologically meaningful dropout events. DDIM captures only half the true sparsity, misrepresenting dropout patterns critical for clustering and imputation \citep{Qiu:2020}. While DDIM-T matches average sparsity, it underperforms quantitatively (Section~\ref{sec:scrna}), as does the domain-specific scDiffusion. For \emph{sparse images}, sparsity mismatch trivializes fake image detection, as generated images lacking authentic sparsity patterns become easily identifiable, compromising robustness in security applications.

\paragraph{Sharpness of generated SBs} As SBs are central to SDD and our model is the first one to simultaneously diffuse the discrete SBs and continuous (dense) inputs, we further examine the distribution of the logits for the SBs in Figure \ref{fig:sparsity_bits_distribution} in Appendix \ref{app:further_sparsity_analysis}. As can be seen, SBs are sharply distributed towards -1 and 1, indicating the model's confidence in predicting these pixel values as zero or non-zero. This indicates that discrete variables can be reliably diffused alongside continuous variables. The last bin with an upper bound of 1.0 shows almost precisely the pixel sparsity of the respective dataset. 

\subsection{Physics: calorimeter image 
generation}
\begin{table}[ht]
\caption{Shown are the results for sparse calorimeter image generation. SDD performs better in preserving the key structural properties of the data—sum of momenta transverse to the beam $P_T$ and invariant mass—and achieves results on par with the domain-specific SARM baseline, demonstrating its competitive performance without relying on domain-specific design.}
\label{tab:sparse_calo_image_generation}
\begin{center}
\begin{sc}
    \begin{tabular}{l cccc}
        \toprule
        & \multicolumn{2}{c}{Signal} &\multicolumn{2}{c}{Background} \\
        Model & $\downarrow$ $P_T$ & $\downarrow$ Mass & $\downarrow$ $P_T$ & $\downarrow$ Mass  \\
        \midrule
        DDPM & 219.30 & 79.19 & 227.00  & 81.43\\
        DDIM & 251.79 & 89.96 & 259.43 & 92.20  \\
        \midrule
        DDPM-T & 23.21 & 11.30 & 27.83 & 12.52 \\
        DDIM-T & 25.36 & 12.16 & 31.28 & 13.81 \\
        \midrule
        SARM D+C & 27.96 & 7.29 & \textbf{12.50} & \textbf{5.39}  \\
        \midrule
        SDD DDPM (ours) &  15.00 & 6.50 & 13.92 & 5.44\\
        SDD DDIM (ours) & \textbf{14.67} & \textbf{6.43}  & 13.01  & 5.41 \\
        \bottomrule
    \end{tabular}
\end{sc}
\end{center}  
\end{table}
%\vspace{-2em}

\paragraph{Quantitative evaluation} The results are summarized in Table \ref{tab:sparse_calo_image_generation}. The Wasserstein Distance $W_P$ between the distributions of the sum of momenta transverse to the beam ($P_T$) and the invariant mass for both signal and background images is substantially larger for DDPM, DDIM, and their thresholded variants (DDPM-T, DDIM-T) compared to SDD DDPM and SDD DDIM, indicating that sparsity-aware SDD generates outputs more closely aligned with the real data. SDD performs exceptionally well on signal images, where the $W_P$ for $P_T$ is worse for SARM. SARM achieves slightly better results for background images, but SDD remains competitive overall. These findings suggest that SDD’s sparsity-aware modeling effectively captures key domain-specific features without relying on extensive prior knowledge, resulting in improved fidelity and realism of generated calorimeter data.

\begin{wrapfigure}{r}{0.5\textwidth}
\centering
\includegraphics[width=\linewidth]{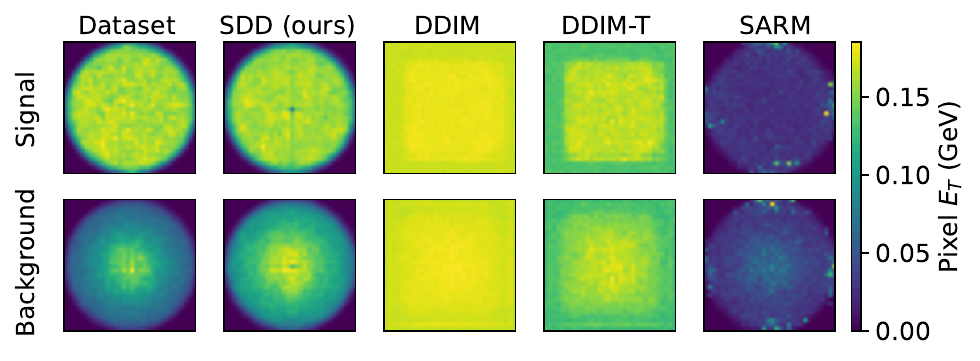}
\caption{Shown are the average calorimeter images for Muon Signal and Muon Background. DDIM and DDIM-T fail to generate realistic data, while SDD succeeds. Linear scale to reveal the signal and background differences.}
\label{fig:ddim_muon_avg}
\end{wrapfigure}

\begin{figure}[t!]
\centering
\begin{subfigure}[b]{0.49\textwidth}
    \centering
    \includegraphics[width=\linewidth]{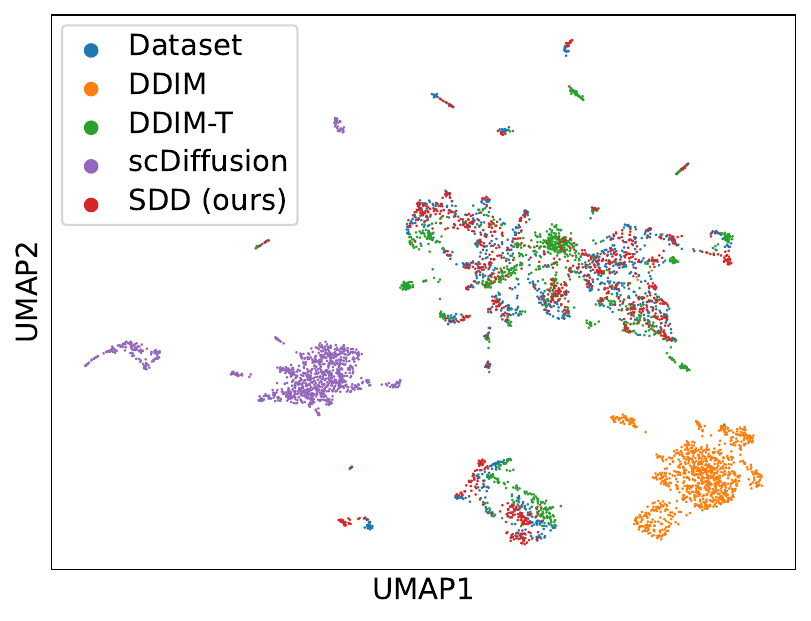}
    \caption{Tabula Muris}
    \label{fig:ddim_tabula_muris_umap}
\end{subfigure}
\begin{subfigure}[b]{0.49\textwidth}
    \centering
    \includegraphics[width=\linewidth]{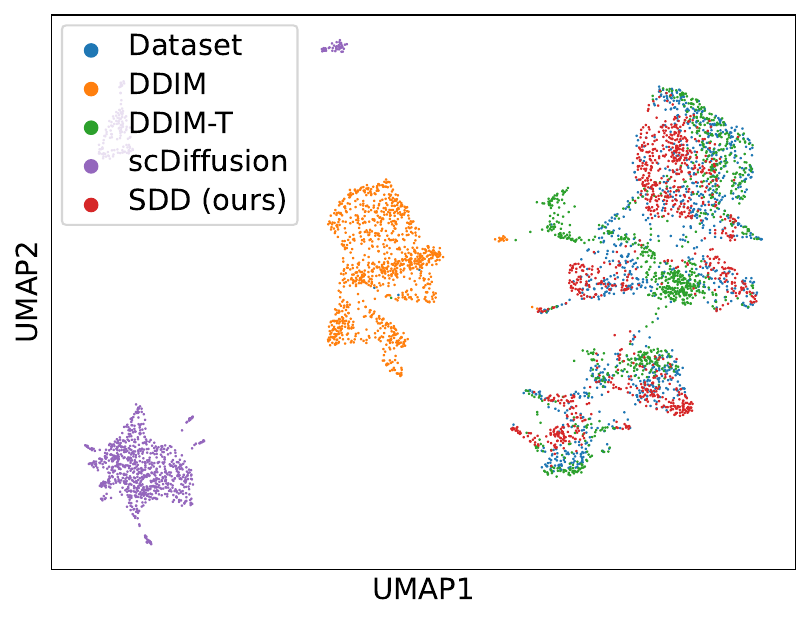}
    \caption{Human Lung Pulmonary Fibrosis}
    \label{fig:ddim_human_lung_pf_umap}
\end{subfigure}
\caption{Shown are two-dimensional UMAPs of Tabula Muris and Human Lung Pulmonary Fibrosis, and the respective DDIM, DDIM-T, and SDD generated samples. DDIM, DDIM-T, and the domain-specific scDiffusion show little to no overlap with the respective dataset while SDD shows significant overlap, demonstrating that SDD more accurately captures the underlying structure and diversity of the real data distributions.}
\label{fig:ddim_scrna}
\end{figure}

\paragraph{Qualitative evaluation} Figure \ref{fig:ddim_muon} shows Muon Signal and Background calorimeter images generated by DDIM (similar DDPM results in Appendix \ref{app:further_physics_results}). DDIM fails to capture clustered energy deposits, while thresholded DDIM-T and SARM produce many isolated single-pixel deposits lacking the clustered deposits of real data. In contrast, SDD accurately reproduces these clusters, consistent with physical interactions. Pixel-wise averages in Figure \ref{fig:ddim_muon_avg} confirm this: DDIM and DDIM-T show almost uniform signals across the image, and SARM underestimates the pixel intensity. For signal images, SDD models the near-uniform energy spread of the calorimeter; for backgrounds, it captures energy near muons from jets. Though DDIM-T and DDPM-T approximate overall sparsity (Section \ref{sec:sparsity_characteristics}), they fail to preserve realistic sparsity patterns and spatial coherence. Similarly, despite SARM’s domain knowledge, it also shows the circular shape of energy deposits but misses the intensity.

\subsection{Biology: scRNA data generation}
\label{sec:scrna}
\paragraph{Quantitative evaluation} Table \ref{tab:sparse_scrna_generation} shows scRNA generation results. Across both Tabula Muris and Human Lung Pulmonary Fibrosis datasets, SDD outperforms standard DDPM/DDIM and their thresholded variants (DDPM-T/DDIM-T) by orders of magnitude. SCC and PCC scores show SDD better preserves gene expression ordering and linearity. MMD distances indicate SDD's distributions are significantly closer to real data, and LISI scores reflect improved mixing between real and generated samples. Notably, SDD outperforms scDiffusion—despite its pretrained autoencoder on a large cell corpus—demonstrating that explicit sparsity modeling captures biological data distributions more effectively than domain-specific pretraining alone.

\begin{table}[t]
\setlength{\tabcolsep}{1mm}
\caption{The results for sparse data generation on scRNA data are shown. SDD outperforms DDPM and DDIM, the thresholded DDPM-T and DDIM-T, and the task-specific scDiffusion in all considered metrics.}
\label{tab:sparse_scrna_generation}
\begin{center}
\begin{sc}
    \begin{tabular}{l c c c c  c c c c}
        \toprule
        & \multicolumn{4}{c}{Tabula Muris} & \multicolumn{4}{c}{Human Lung PF} \\
        Model & $\uparrow$ SCC & $\uparrow$ PCC & $\downarrow$ MMD & $\uparrow$ LISI & $\uparrow$ SCC & $\uparrow$ PCC & $\downarrow$ MMD & $\uparrow$ LISI \\
        \midrule
        DDPM & 0.49 & 0.72 & 3.58 & 0.00 & 0.31 & 0.85 & 3.35 & 0.00 \\
        DDIM & 0.50 & 0.74 & 3.62 & 0.00 & 0.30 & 0.87 & 3.34 & 0.00 \\
        \midrule
        DDPM-T & 0.53 & 0.68 & 0.35 & 0.01 & 0.31 & 0.82 & 0.69 & 0.00 \\
        DDIM-T & 0.56 & 0.73 & 0.34 & 0.01 & 0.28 & 0.85 & 0.65 & 0.00 \\
        \midrule
        scDiffusion & 0.71 & 0.71 & 1.54 & 0.00 & 0.77 & 0.79 & 1.02 & 0.00 \\
        \midrule
        SDD DDPM (ours) & 0.96 & 0.95 & \textbf{0.12} & 0.21 & \textbf{0.95} & \textbf{0.99} & \textbf{0.18} & \textbf{0.08} \\
        SDD DDIM (ours) & \textbf{0.97} & \textbf{0.97} & \textbf{0.12} & \textbf{0.23} & \textbf{0.95} & \textbf{0.99} & \textbf{0.18} & \textbf{0.08} \\
        \bottomrule
    \end{tabular}
\end{sc}
\end{center}  
\end{table}

\paragraph{Qualitative evaluation} In our qualitative evaluation for scRNA data, we provide UMAPs \citep{Mcinnes:2018} for Tabula Muris and Human Lung Pulmonary Fibrosis in Figure \ref{fig:ddim_scrna} for DDIM, with further analysis showing similar results for DDPM in Appendix \ref{app:further_biology_results}. Both subfigures in Figure \ref{fig:ddim_scrna} show almost no overlap between real data and DDIM-generated data. For DDIM-T, the generated cell cluster shifts closer to the real cells but exhibits little diversity. The domain-specific scDiffusion model also demonstrates little to no overlap with real cells, indicating it fails to capture the true biological structure despite its pretraining on a large cell corpus. In contrast, SDD shows significant overlap between real cells and those sampled from SDD, further underlining the high quality and biological fidelity of the generated cells.

\subsection{Vision: sparse image generation}
\label{sec:sparse_image_generation}
\paragraph{Quantitative evaluation} For sparse image generation, we summarize our results in Table~\ref{tab:sparse_image_generation} in Appendix \ref{app:sparse_image_generation}. We observe similar FID scores for all settings considered. However, the validity of FID for evaluating sparse images is questionable, as the standard Inception network used to compute FID was trained only on dense images from ImageNet \citep{Deng:2009}, which may limit its effectiveness for sparse data. Therefore, we report these values only for completeness. 

\paragraph{Qualitative evaluation} The quality of generated images on Fashion-MNIST and MNIST (see Figures~\ref{fig:fashion_mnist_sparsity} and \ref{fig:mnist_sparsity} in Appendix~\ref{app:sparse_image_generation}) is largely indistinguishable across DDIM, DDPM, their thresholded variants (DDIM-T, DDPM-T), and SDD using DDPM and DDIM. Visually, all models produce plausible samples. However, a closer inspection of the sparsity patterns reveals meaningful differences. In particular, SDD more faithfully reproduces the sparsity structure of the real data. For Fashion-MNIST, thresholded models like DDIM-T and DDPM-T tend to suppress fine-grained noisy regions near the edges. Similarly, in MNIST, the digits generated by DDIM-T and DDPM-T are generally narrower than those in real data. These subtle differences highlight that, while thresholding may yield visually appealing outputs, it often fails to preserve key structural properties of the data, particularly in challenging regions where sparsity transitions are critical. SDD, by contrast, maintains these characteristics more reliably, capturing both visual fidelity and underlying data distribution properties by explicitly modeling the sparsity.

\section{Conclusion}
\label{sec:conclusion}
We introduce SDD, a physically-grounded method for generating sparse scientific data using DMs. By representing sparsity through Sparsity Bits—discrete variables diffused alongside continuous values—SDD explicitly encodes the physical structure of sparse data. We  demonstrate high fidelity across physics and biology applications, with additional validation on computer vision tasks. SDD achieves superior sample quality compared to standard DMs and their thresholded variants, matching the domain-specific SARM in particle physics and surpassing the biologically pre-trained scDiffusion for scRNA data. This demonstrates that explicit sparsity modeling is more effective than approaches relying solely on domain knowledge or pretraining. The underlying concept of Sparsity Bits can be integrated into other generative models, including GANs \citep{Goodfellow:2020}, VAEs \citep{Kingma:2014}, and Normalizing Flows \citep{Rezende:2015}, opening new research directions for physically-grounded generative modeling in scientific simulation.

\begin{ack}
This work was conducted within the initiative AI-Care by the Carl-Zeiss Stiftung. Part of this work was conducted within the DFG Research Unit FOR 5359 on Deep Learning on Sparse Chemical Process Data (BU 4042/2-1, KL 2698/6-1, and KL 2698/7-1). MK and SF further acknowledge support by the DFG TRR 375 (ID 511263698), the DFG SPP 2298 (KL 2698/5-2), and the DFG SPP 2331 (FE 2282/1-2, FE 2282/6-1, and KL 2698/11-1). Additional support was provided by the BMBF award 01|S2407A.
\end{ack}

\bibliography{paper}
\bibliographystyle{plainnat}

%%%%%%%%%%%%%%%%%%%%%%%%%%%%%%%%%%%%%%%%%%%%%%%%%%%%%%%%%%%%

\appendix
\newpage
\section{Training and sampling algorithms}
\label{app:train_sample}
In this section, we present SDD's training procedure in Algorithm \ref{alg:training} and the sampling in Algorithm \ref{alg:sampling}.

\lstset{
    escapeinside={(*@}{@*)},  % Allow escaping to LaTeX code
    columns=flexible,
    basicstyle=\ttfamily,
    language=Python,
%    numbers=left
}

\begin{algorithm}[ht]
\caption{SDD training algorithm.}
\begin{lstlisting}[]
def train_loss(x):
  # Create & concat sparsity bits
  x_sb = (x != 0).float()
  x_sb = (x_sb * 2 - 1)
  x = (x * 2 - 1)
  x = cat((x, x_sb), dim=1)
    
  # Forward diffusion steps
  t = uniform(0, 1)
  eps = normal(mean=0, std=1)
  x_t = sqrt(alpha(t)) * x + (1 - sqrt(alpha(t)) * eps

  # Compute self-cond estimate. 
  x_0 = zeros_like(x_t)
  if uniform(0, 1) > 0.5:
    x_0 = net(cat([x_t, x_0], -1), t)
    x_0 = stop_gradient(x_0) 
    
  # Predict and compute loss. 
  x_0 = net(cat([x_t, x_0], -1), t) 
  l2_loss = (x_0[:, :sb] - x[:, :sb]) ** 2
  ce_loss = cross_entropy_loss(x_0[:, sb:] - x[:, sb:])
  return l2_loss.mean() + ce_loss.mean()
\end{lstlisting}
\label{alg:training}
\end{algorithm}

\begin{algorithm}[ht]
\caption{SDD sampling algorithm.}
\begin{lstlisting}[language=Python]
def sample(steps, interm=False):
  x_t = randn(mean=0, std=1)
  x_pred = zeros_like(x_t)
    
  for step in range(steps):
    # Get time for current & next states. 
    t_now = 1 - step / steps 
    t_next = max(1 - (step + 1)/steps, 0)

    # Predict x_0. 
    x_pred = net(cat([x_t, x_pred], -1), t_now)
        
    # Estimate x at t_next. 
    x_t = ddim_or_ddpm_step(x_t, x_pred, t_now, t_next)

  # Return sparsified data point
  x_pred.clamp_(-1,1)
  x_dense = (x_pred[:, :sb] + 1) / 2
  x_sb = (x_pred[:, sb:] > 0).float()
  x_sparse = x_dense * x_sb
  return x_sparse
\end{lstlisting}
\label{alg:sampling}
\end{algorithm}

\section{DM architecture \& training}
\label{app:dm_architecture}
We use a CNN-based U-Net architecture \citep{Ho:2020, Ronneberger:2015, Nichol:2022} for all DMs on calorimeter and sparse image data, with a base channel size of 256, three stages, two residual blocks per stage, and 37M parameters. For scRNA data, which lacks spatial structure, we adopt a skip-connected MLP following scDiffusion \citep{Luo:2024} with 5M parameters. Parameter counts and runtime overhead are detailed in Appendix \ref{app:runtime}. We use Adam \citep{Kingma:2015} with a constant learning rate of 0.0002 and batch size 256. Models are trained for 300K steps, using an exponential moving average of parameters (decay 0.9999). Sampling for DDIM and DDPM uses 1,000 steps.

\section{Compute resources and runtime analysis}
\label{app:runtime}
Our experiments were executed on a NVIDIA DGX-A100 server with eight A100 (40GB) GPUs, an AMD Epyc 7742 CPU with 64 cores and 2TB of main memory. The server runs NVIDIA DGX Server Version 7.1.0 (GNU/Linux 6.8.0-60-generic x86\_64) and has CUDA 12.9 and torch 2.7.0 installed. Due to the large scale and high computational cost of the deployed DMs, we were only able to run each experiment once. Each run requires substantial GPU resources and time making multiple runs or significance testing infeasible. Additionally, sampling is particularly expensive, further limiting repetition. While we acknowledge the importance of variance analysis, these practical constraints necessitated a single-run evaluation.

SDD enforces sparsity in the generated data and does not exploit the data sparsity to make the computations more efficient. Depending on the underlying U-Net architecture, we have more trainable parameters. We use CNNs with a total of 37M parameters for the calorimeter and the sparse images. However, for SDD, we observe an increase of only 0.07\% in the number of parameters. For sparse data generation of scRNA data with an MLP with a total of 5M parameters, we observe an increase of 38.37\% in the number of parameters. The increase in the number of parameters is much higher, as we have to double the size of the input and output layers. At the same time, we only need to double the number of input and output channels for the CNN-based U-Net for images. 

We summarize the training runtime of DDPM and SDD in Table \ref{tab:runtimes}. We omit DDIM as the DDPM and DDIM variants have the same training procedure.  We measured runtimes per training step on one A100 (40GB) that show minimal overhead for SDD compared to DDPM, as additional computations are efficiently parallelized.

\begin{table}[ht]
\caption{Shown are the runtimes of DDPM compared to SDD. The overhead for SDD is minimal.}
\label{tab:runtimes}
\centering
\begin{sc}
\begin{tabular}{l  l  c  c}
    \toprule
    UNet & Dataset &  DDPM & SDD\\
    \midrule
    CNN & MNIST & 0.57 s & 0.56 s\\
    CNN & Fashion-MNIST & 0.50 s & 0.51 s \\
    \midrule
    CNN & Muon Signal & 0.56 s & 0.56 s \\
    CNN & Muon Background & 0.59 s & 0.58 s \\
    \midrule
    MLP & Tabula Muris & 0.02 s & 0.02 s \\
    MLP & Human Lung PF & 0.02 s & 0.02 s \\
    \bottomrule
\end{tabular}

\end{sc}
\end{table}

\section{Further evaluating the recovery of ground truth sparsity patterns}
\label{app:further_sparsity_analysis}

\begin{figure}[ht]
\includegraphics[width=\columnwidth]{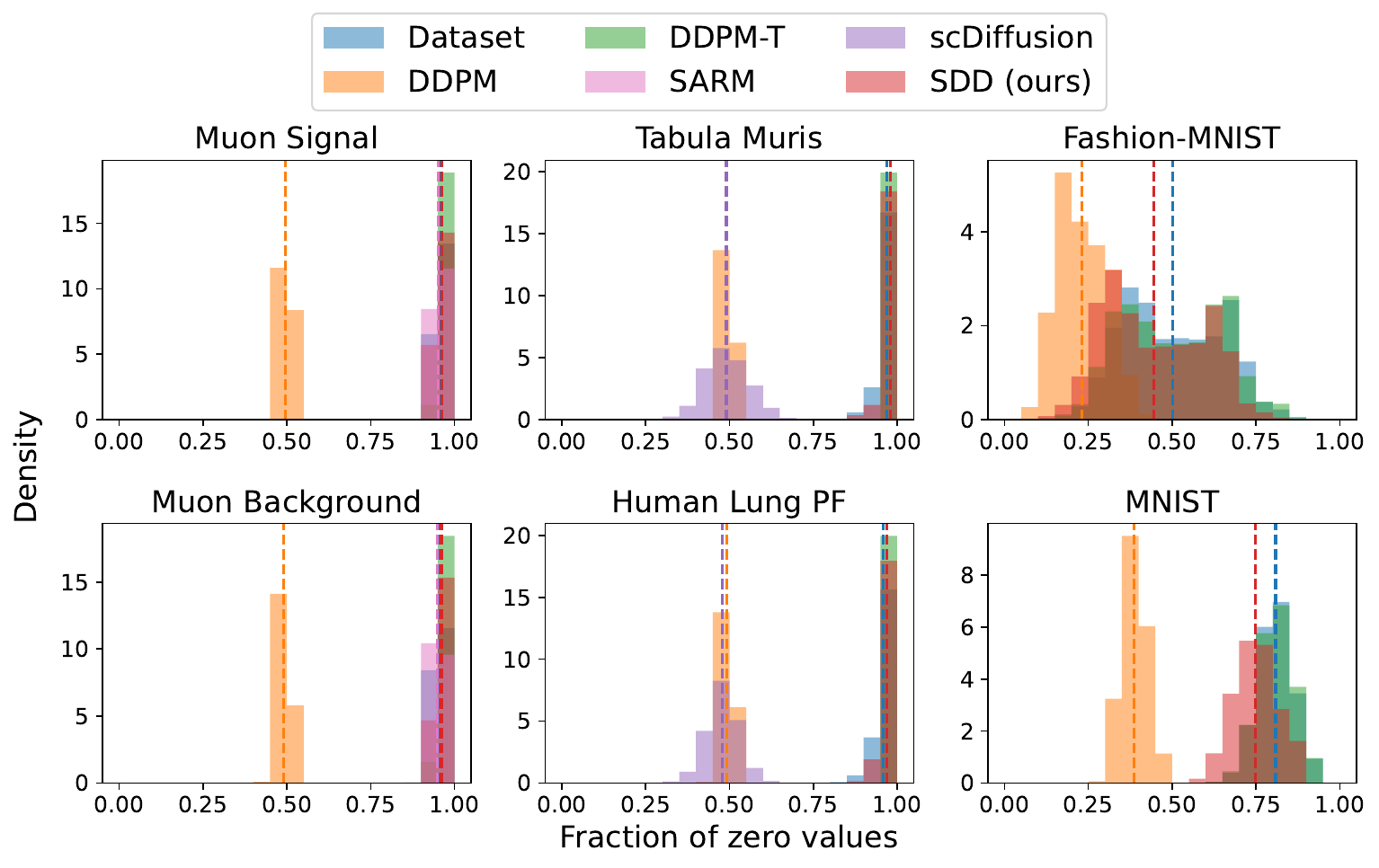}
\caption{Sparsity distribution of real and generated data. Histograms (20 bins) and average sparsity levels (dashed lines) compare real data to samples from DDPM, DDPM-T, SARM, scDiffusion, and SDD. DDPM and scDiffusion underestimate sparsity; DDPM-T matches the average but lacks diversity and overshoots at high sparsity. SARM also underestimates sparsity, while SDD accurately matches both average value and distribution.}
\label{fig:ddpm_sparsity_distribution}
\end{figure}
\paragraph{How well do models match ground-truth sparsity?} 

We present the sparsity distribution of samples generated by DDPM in Figure~\ref{fig:ddpm_sparsity_distribution}, complementing the main text's analysis of DDIM (Figure~\ref{fig:ddim_sparsity_distribution}). The trends observed largely mirror those of DDIM. Specifically, DDPM also tends to underestimate sparsity across all datasets. Similar to DDIM-T, DDPM-T produces samples with an exaggerated upper tail in the sparsity distribution and a compressed lower end, particularly noticeable in the sparse image generation datasets. SDD remains the most faithful to the real data’s sparsity profile: it mildly overshoots in the physics datasets (Muon Signal and Muon Background) and does so more substantially in the scRNA datasets (Tabula Muris and Human Lung Pulmonary Fibrosis), though still less severely than DDPM-T. On the vision datasets (Fashion-MNIST and MNIST), SDD continues to generate a wider and more accurate sparsity distribution.

\begin{figure}[ht]
\centering
    \includegraphics[width=\columnwidth]{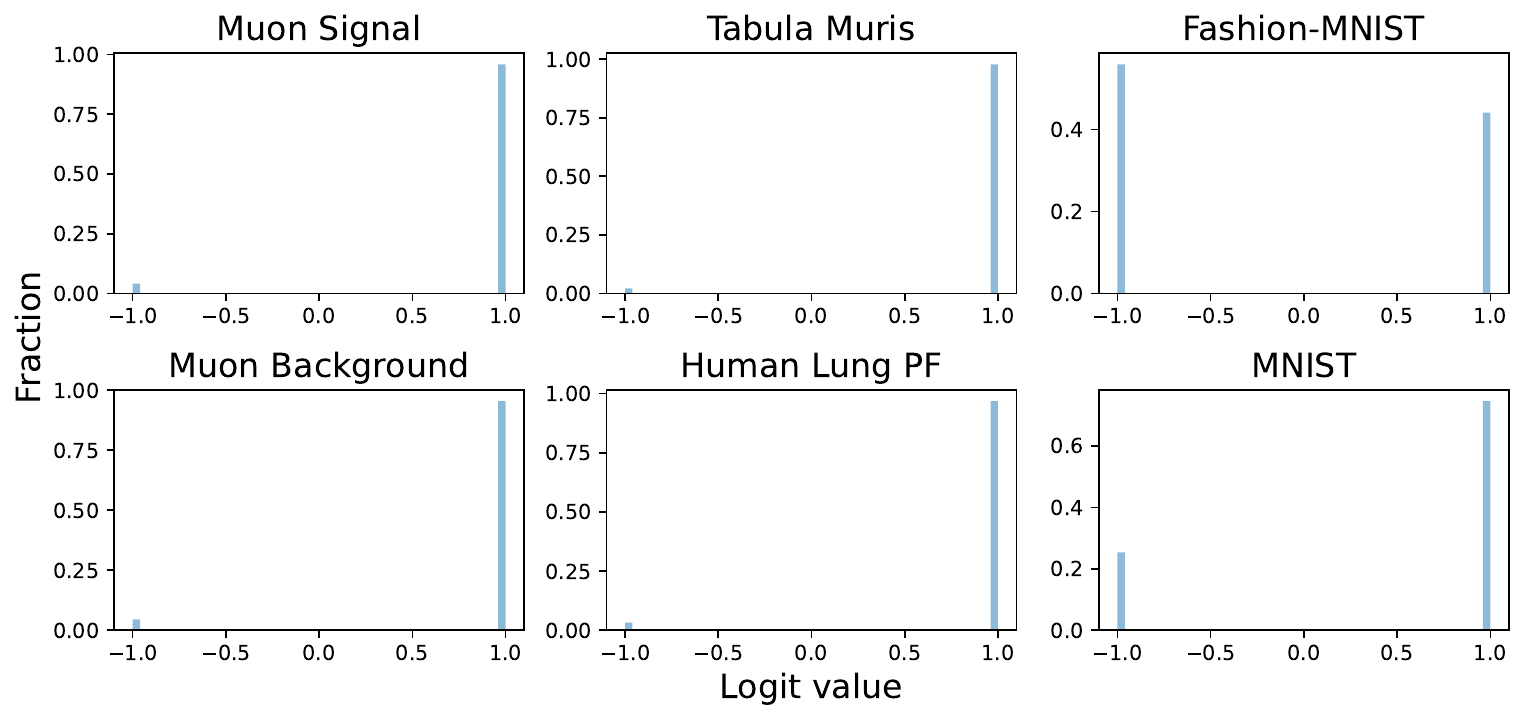}
\caption{The histogram distribution shows 50 bins of the logits for sparsity bits for datapoints using SDD. The sparsity bits are extremely concentrated towards -1 and 1, which shows that discrete variables can be reliably diffused alongside continuous variables.}
\label{fig:sparsity_bits_distribution}
\end{figure}

\paragraph{Sharpness of generated SBs}
Figure~\ref{fig:sparsity_bits_distribution} illustrates the distribution of the sparsity bit (SB) logits across datasets. The histograms show that the logits are sharply concentrated near -1 and 1, indicating high model confidence when classifying each dimension as corresponding to a zero or non-zero pixel. This supports the feasibility of jointly diffusing discrete and continuous components. Moreover, the final bin (with upper bound 1.0) closely aligns with the true sparsity levels of each dataset: 95.2\% for Muon Signal, 97.9\% for Tabula Muris, 50.2\% for Fashion-MNIST, 95.2\% for Muon Background, 95.9\% for Human Lung Pulmonary Fibrosis, and 80.9\% for MNIST.

\section{Further results for particle physics: calorimeter image generation}
\label{app:further_physics_results}
\begin{figure}[ht]
\centering
\begin{subfigure}[b]{0.49\columnwidth}
    \centering
    \includegraphics[width=\columnwidth]{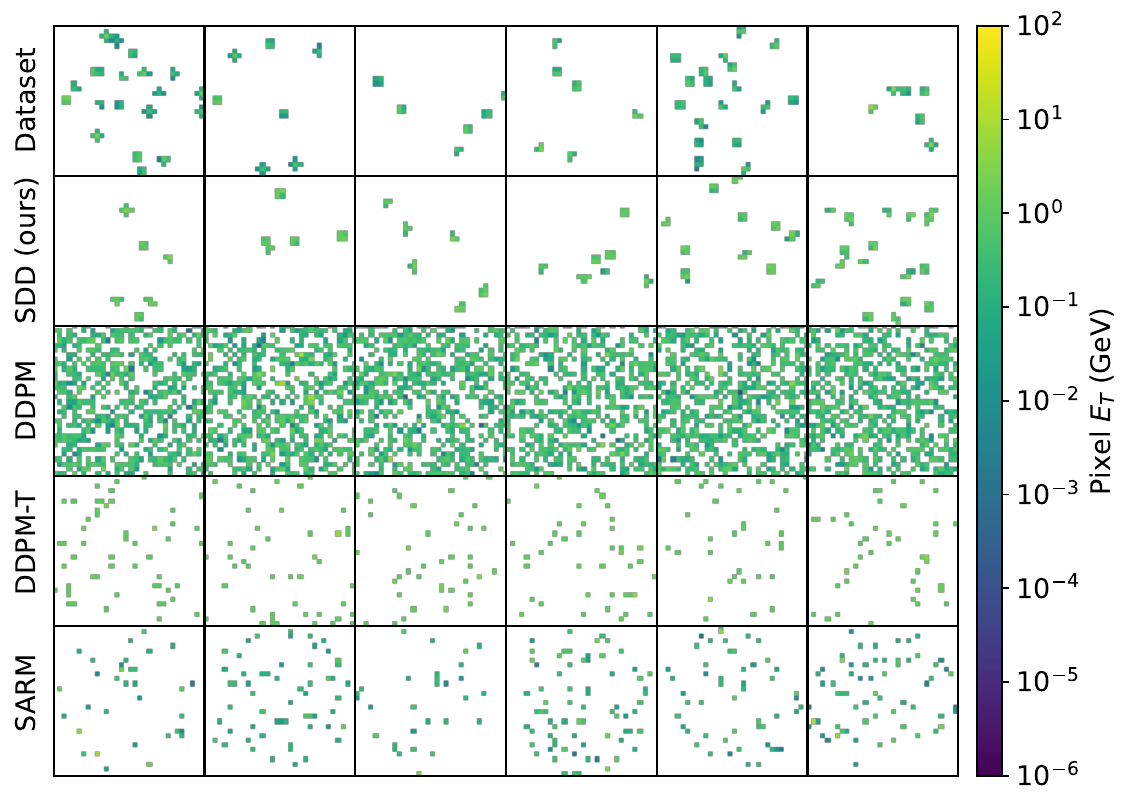}
    \caption{Muon Signal}
    \label{fig:ddpm_muon_signal}
\end{subfigure}
\begin{subfigure}[b]{0.49\columnwidth}
    \centering
    \includegraphics[width=\columnwidth]{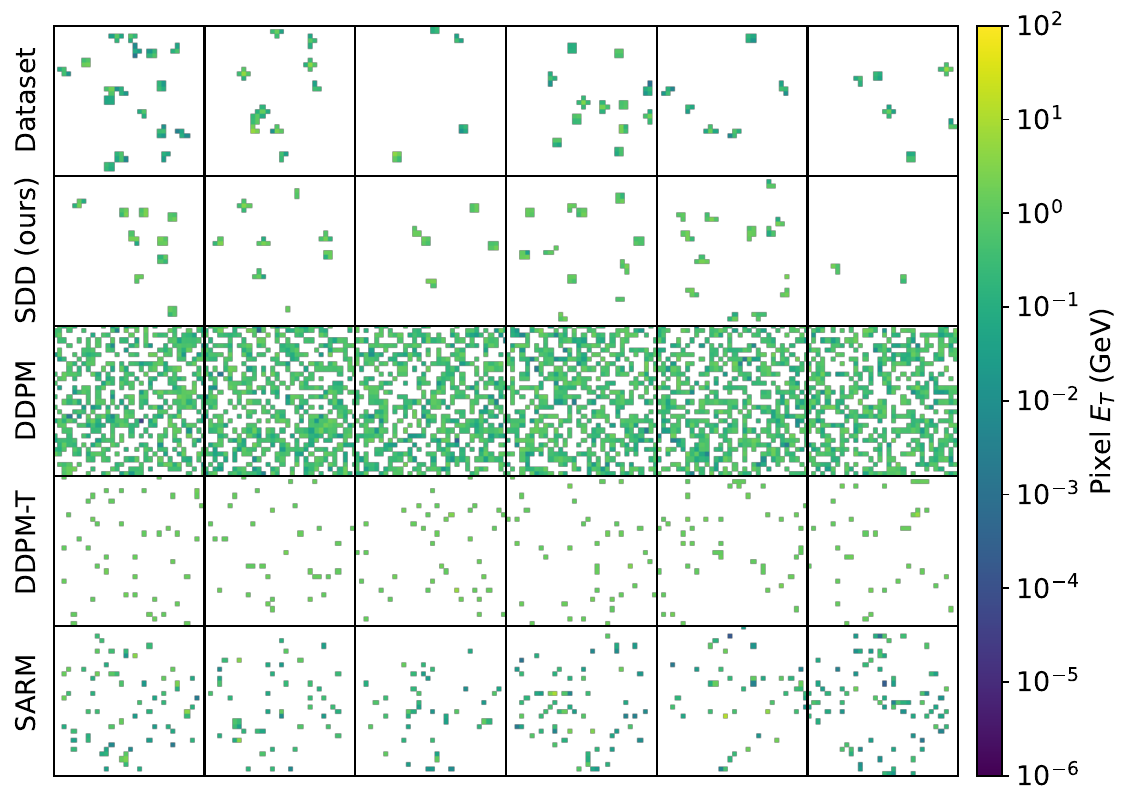}
    \caption{Muon Background}
    \label{fig:ddpm_muon_background}
\end{subfigure}
\caption{Calorimeter images from the muon isolation study: signal images (top) and background images (bottom). Rows show samples from (1) real data, (2) SDD (ours), (3) DDPM, (4) DDPM with post-hoc thresholding to match dataset sparsity (DDPM-T), and (5) the domain-specific SARM baseline. Pixel intensity (GeV) visualizes energy deposition per cell (white=zero). SDD uniquely recovers the distinct, sparse, and clustered patterns characteristic of real data, whereas DDPM completely misses the sparsity, and DDPM-T and SARM show unrealistic isolated energy deposits.}
\label{fig:ddpm_muon}
\end{figure}
Figure~\ref{fig:ddpm_muon} displays representative calorimeter images generated by DDPM for two distinct cases: isolated muons (signal) and muons produced in association with jets (background). These qualitative results reveal behaviors similar to those observed for DDIM in Figure~\ref{fig:ddim_muon}, illustrating fundamental limitations inherent in standard diffusion models when applied to sparse data. Specifically, DDPM-generated images fail to faithfully reproduce the localized and clustered energy depositions that characterize true muon interactions in the calorimeter. Instead, the energy distribution appears more diffuse and lacks the sharp, spatially coherent patterns critical for meaningful physics interpretation.

The thresholded variant, DDPM-T, which applies a post-hoc procedure to enforce dataset-level sparsity by zeroing out low-intensity pixels, mitigates some of these issues by producing more spatially isolated activations. However, these tend to manifest largely as single-pixel hits scattered across the calorimeter grid rather than physically plausible clusters, indicating that simple thresholding does not recover the underlying complex sparsity structure or spatial correlations.

In stark contrast, SDD-DDPM generates highly realistic calorimeter images that significantly improve these shortcomings. By explicitly modeling sparsity through Sparsity Bits and jointly diffusing continuous and discrete latent representations, SDD-DDPM reproduces the expected clustered energy patterns with remarkable fidelity. This confirms that incorporating sparsity awareness directly into the generative process enables the model to capture the intricate spatial dependencies and zero patterns intrinsic to the data.

\begin{figure}[ht]
\centering
\includegraphics[width=0.8\columnwidth]{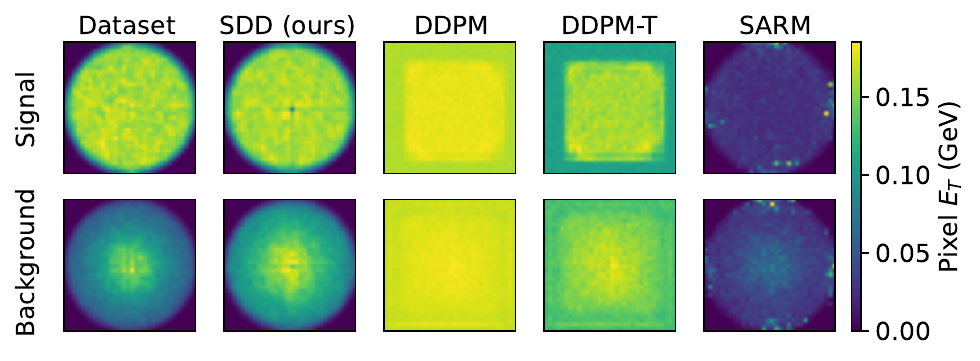}
\caption{Averaged calorimeter images for Muon Signal and Background. SDD-generated averages closely match the actual data’s spatial structure and sparsity. DDPM and DDPM-T produce almost uniform images, while the domain-specific SARM baseline fails to capture the actual intensity.}
\label{fig:ddpm_muon_avg}
\end{figure}
Complementary evidence is provided by the pixel-wise average calorimeter images illustrated in Figure~\ref{fig:ddpm_muon_avg}. These averages further underscore the qualitative differences: both DDPM and DDPM-T exhibit relatively uniform, blurred activations across the image, indicative of non-specific and diffuse energy distributions that do not differentiate between zero and non-zero values effectively. By contrast, the averages constructed from SDD-DDPM samples reveal well-defined and physically meaningful spatial structures. For signal images, SDD-DDPM captures the characteristic uniform cylindrical energy distribution observed in isolated muon events. In contrast, in the background case, the energy is correctly concentrated near the muon’s expected location rather than dispersed randomly. These distinct and interpretable patterns closely mirror the true physical phenomena measured by the calorimeter, substantiating the advantage of SDD in sparse data generation.

\begin{figure}[ht!]
\centering
\begin{subfigure}[b]{0.49\columnwidth}
    \centering
    \includegraphics[width=\columnwidth]{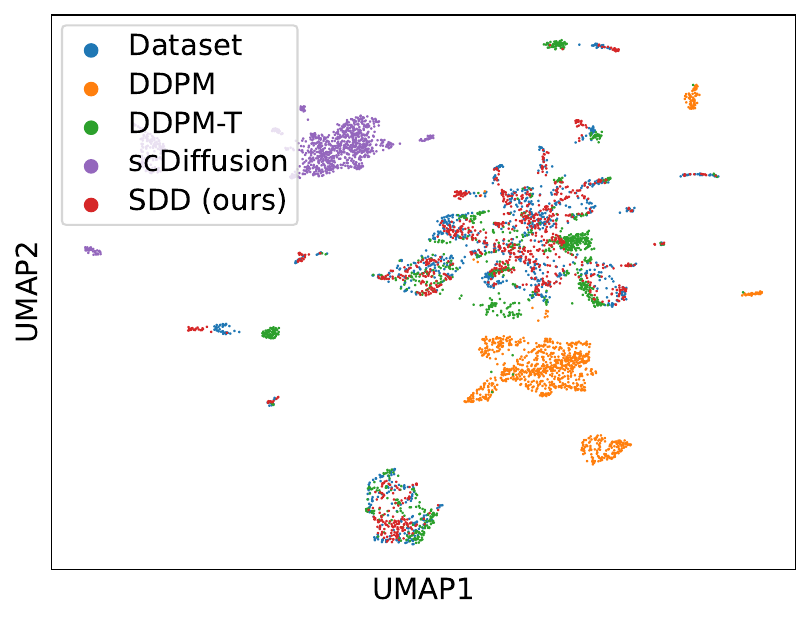}
    \caption{Tabula Muris}
    \label{fig:ddpm_tabula_muris_umap}
\end{subfigure}
\begin{subfigure}[b]{0.49\columnwidth}
    \centering
    \includegraphics[width=\columnwidth]{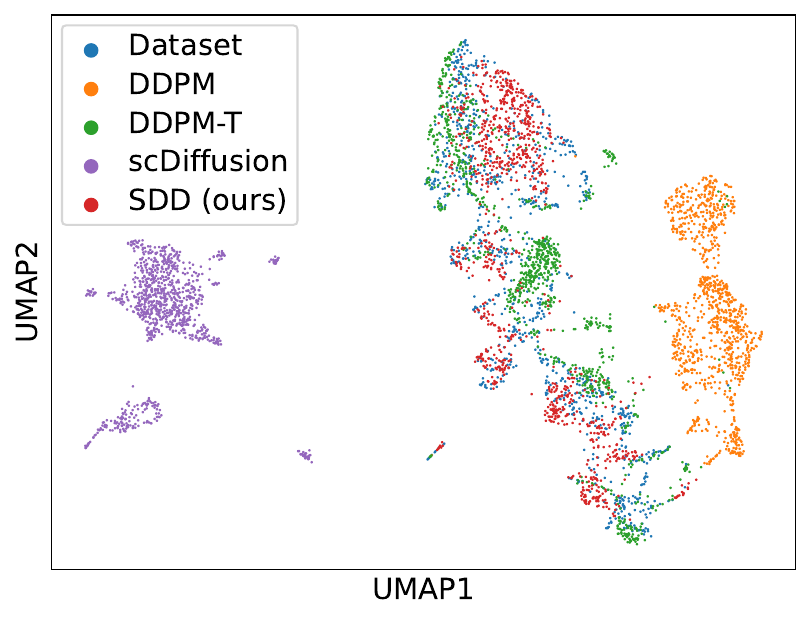}
    \caption{Human Lung Pulmonary Fibrosis}
    \label{fig:ddpm_human_lung_pf_umap}
\end{subfigure}
\caption{Shown are two-dimensional UMAPs of Tabula Muris and Human Lung Pulmonary Fibrosis, and the respective DDPM, DDPM-T, and SDD generated samples. DDPM, DDPM-T, and the domain-specific scDiffusion show little to no overlap with the respective dataset while SDD shows significant overlap, demonstrating that SDD more accurately captures the underlying structure and diversity of the real data distributions.}
\label{fig:ddpm_scrna}
\end{figure}

\section{Further results for biology: scRNA data generation}
\label{app:further_biology_results}
Figure~\ref{fig:ddpm_scrna} shows UMAP visualizations for the Tabula Muris and Human Lung Pulmonary Fibrosis datasets generated using DDPM. The trends mirror those observed with DDIM (Figure~\ref{fig:ddim_scrna}). Specifically, DDPM-generated samples show minimal overlap with the real cell populations, indicating limited biological realism. The thresholded variant, DDPM-T, moves the generated cluster closer to the real data manifold but at the cost of reduced diversity. In contrast, SDD-DDPM samples exhibit substantial overlap with the true cell populations, capturing both structure and diversity more faithfully. These results confirm that the benefits of sparsity-aware diffusion extend to DDPM-based models in the biological domain.

\section{Further Results for Sparse Image Generation}

\begin{table}[ht]
\caption{Shown are the FID scores for sparse image generation. The results are similar for all considered settings.}
\label{tab:sparse_image_generation}

\center
\begin{sc}
    \begin{tabular}{l c c}
        \toprule
        & Fashion-MNIST & MNIST \\
        \midrule
        DDPM & 25.62 & 23.35 \\
        DDIM  & 24.11 & 23.68 \\
        \midrule
        DDPM-T & 25.83 & 23.82 \\
        DDIM-T & 25.72 & 24.04 \\
        \midrule
        SDD DDPM (ours)  & 27.81 & 25.37\\
        SDD DDIM (ours)  & 26.37 & 25.69\\
        \bottomrule  
    \end{tabular}
\end{sc}
\end{table}

\label{app:sparse_image_generation}
\begin{figure}[t]
    \centering
    \begin{subfigure}[b]{\columnwidth}
        \centering
        \includegraphics[width=\columnwidth]{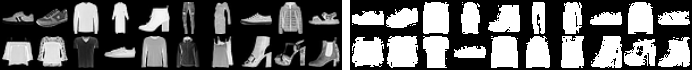}
        \caption{Dataset}
        \label{fig:sparsity_fashion_mnist}
    \end{subfigure}
    \begin{subfigure}[b]{\columnwidth}
        \centering
        \includegraphics[width=\columnwidth]{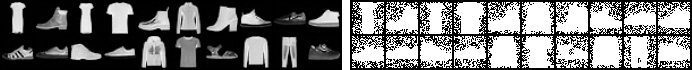}
        \caption{DDPM}
        \label{fig:sparsity_ddpm_fashion_mnist}
    \end{subfigure}
    \begin{subfigure}[b]{\columnwidth}
        \centering
        \includegraphics[width=\columnwidth]{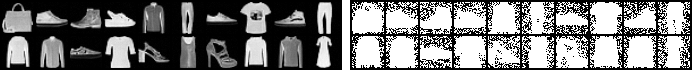}
        \caption{DDIM}
        \label{fig:sparsity_ddim_fashion_mnist}
    \end{subfigure}
    \begin{subfigure}[b]{\columnwidth}
        \centering
        \includegraphics[width=\columnwidth]{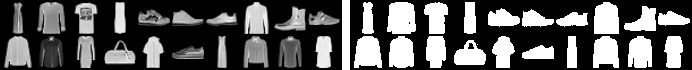}
        \caption{DDPM-T}
        \label{fig:sparsity_ddpm_t_fashion_mnist}
    \end{subfigure}
    \begin{subfigure}[b]{\columnwidth}
        \centering
        \includegraphics[width=\columnwidth]{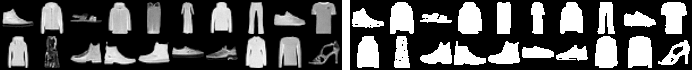}
        \caption{DDIM-T}
        \label{fig:sparsity_ddim_t_fashion_mnist}
    \end{subfigure}
    \begin{subfigure}[b]{\columnwidth}
        \centering
        \includegraphics[width=\columnwidth]{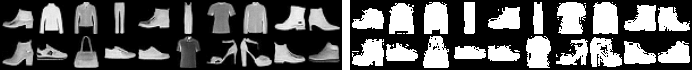}
        \caption{SDD DDPM (ours)}
        \label{fig:sparsity_sdd_ddpm_fashion_mnist}
    \end{subfigure}
    \begin{subfigure}[b]{\columnwidth}
        \centering
        \includegraphics[width=\columnwidth]{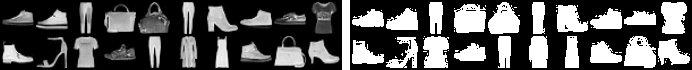}
        \caption{SDD DDIM (ours)}
        \label{fig:sparsity_sdd_ddim_fashion_mnist}
    \end{subfigure}
\caption{Shown are, from top to bottom: Fashion-MNIST images sampled from the dataset, DDPM and DDIM sampled images, thresholded DDIM and DDPM sampled images (DDPM-T, DDIM-T), and SDD (DDPM, DDIM) sampled images. The first column contains the samples, and the second contains the respective sparsity information. Despite highly visually similar images, DDIM and DDPM fail to reflect the sparsity, while the thresholded variants DDIM-T and DDPM-T miss fine-grained details on the edges. The proposed SDD more faithfully reflects these fine-grained details.}
\label{fig:fashion_mnist_sparsity}
\end{figure}

\begin{figure}[t]
    \centering
    \begin{subfigure}[b]{\columnwidth}
        \centering
        \includegraphics[width=\columnwidth]{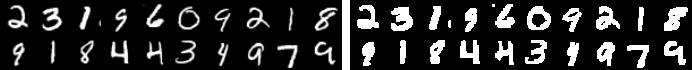}
        \caption{Dataset}
        \label{fig:sparsity_mnist}
    \end{subfigure}
    \begin{subfigure}[b]{\columnwidth}
        \centering
        \includegraphics[width=\columnwidth]{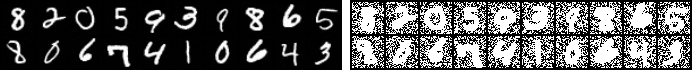}
        \caption{DDPM}
        \label{fig:sparsity_ddpm_mnist}
    \end{subfigure}
    \begin{subfigure}[b]{\columnwidth}
        \centering
        \includegraphics[width=\columnwidth]{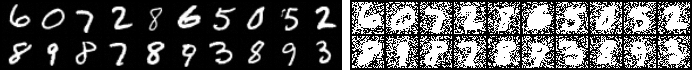}
        \caption{DDIM}
        \label{fig:sparsity_ddim_mnist}
    \end{subfigure}
    \begin{subfigure}[b]{\columnwidth}
        \centering
        \includegraphics[width=\columnwidth]{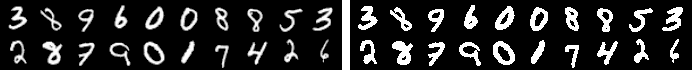}
        \caption{DDPM-T}
        \label{fig:sparsity_ddpm_t_mnist}
    \end{subfigure}
    \begin{subfigure}[b]{\columnwidth}
        \centering
        \includegraphics[width=\columnwidth]{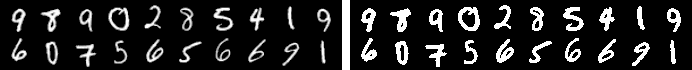}
        \caption{DDIM-T}
        \label{fig:sparsity_ddim_t_mnist}
    \end{subfigure}
    \begin{subfigure}[b]{\columnwidth}
        \centering
        \includegraphics[width=\columnwidth]{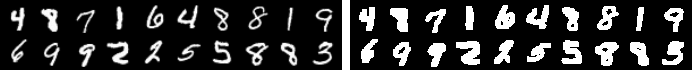}
        \caption{SDD DDPM (ours)}
        \label{fig:sparsity_sdd_ddpm_mnist}
     \end{subfigure}
    \begin{subfigure}[b]{\columnwidth}
        \centering
        \includegraphics[width=\columnwidth]{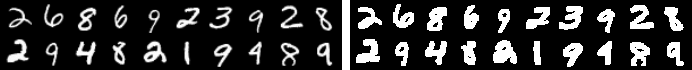}
        \caption{SDD DDIM (ours)}
        \label{fig:sparsity_sdd_ddim_mnist}
     \end{subfigure}
\caption{Shown are, from top to bottom: MNIST images sampled from the dataset, DDPM and DDIM sampled images, thresholded DDIM and DDPM sampled images (DDPM-T, DDIM-T), and SDD (DDPM, DDIM) sampled images. The first column contains the samples, and the second contains the respective sparsity information. Despite highly visually similar images, DDIM and DDPM fail to reflect the sparsity, while the thresholded variants DDIM-T and DDPM-T miss fine-grained details on the edges, resulting in narrower digits (and the respective sparsity information). The proposed SDD more faithfully reflects these fine-grained details.}
\label{fig:mnist_sparsity}
\end{figure}

As discussed in Section \ref{sec:sparse_image_generation}, the visual quality of generated images for Fashion-MNIST and MNIST is largely comparable across all methods—DDIM, DDPM, their thresholded counterparts (DDIM-T and DDPM-T), and SDD. To complement these observations, Figures~\ref{fig:fashion_mnist_sparsity} and \ref{fig:mnist_sparsity} provide detailed visualizations of the corresponding sparsity patterns. While DDIM, DDPM, and their thresholded versions produce images that appear visually plausible, closer examination reveals discrepancies in how well they capture the underlying sparsity structure of the datasets.

In particular, thresholded models such as DDIM-T and DDPM-T often fail to reproduce the fine-grained noise and soft sparsity transitions present in real data. For Fashion-MNIST, these models tend to eliminate subtle pixel activations near the edges, leading to overly clean object contours. In MNIST, the digits generated by DDIM-T and DDPM-T are noticeably narrower, lacking some of the variability found in true samples. These limitations underscore the fact that simple thresholding, while improving sparsity alignment on average, can suppress important structural variation in the data.

By contrast, SDD not only maintains competitive visual quality but also more accurately reproduces the sparsity distributions and edge-region behavior of real images. This highlights the advantage of explicitly modeling sparsity rather than relying on post-hoc thresholding to enforce it.

\end{document}